\documentclass[10pt, conference]{IEEEtran}
\usepackage[utf8]{inputenc}
\usepackage{csquotes}
\usepackage{amsmath,amssymb,amsfonts}
\usepackage{graphicx}
\usepackage{textcomp}
\usepackage{xcolor}
\usepackage{multirow}
\usepackage{dcolumn}

\usepackage{threeparttable}
\usepackage{booktabs}

\usepackage[ruled,vlined]{algorithm2e}

\usepackage{tikz}
\usetikzlibrary{arrows.meta}
\usepackage{listofitems} 

\author{
\IEEEauthorblockN{
Adam Orucu\IEEEauthorrefmark{2}\IEEEauthorrefmark{3},
Marcus Medhage\IEEEauthorrefmark{1},
Farnaz Moradi\IEEEauthorrefmark{2},
Andreas Johnsson\IEEEauthorrefmark{2}\IEEEauthorrefmark{4}, and
Sarunas Girdzijauskas\IEEEauthorrefmark{3}
}

\IEEEauthorblockA{\IEEEauthorrefmark{2} Ericsson Research, Stockholm, Sweden }
\IEEEauthorblockA{\IEEEauthorrefmark{3} KTH Royal Institute of Technology, Stockholm, Sweden}
\IEEEauthorblockA{\IEEEauthorrefmark{4} Uppsala University, Uppsala, Sweden }
}

\usepackage[nolist]{acronym}

\newcommand{\method}{TabGNS}

\begin{document}
\begin{acronym}
\acro{NAS}{Neural Architecture Search}
\acro{ML}{Machine Learning}
\acro{AI}{Artificial Intelligence}
\acro{DL}{Deep Learning}
\acro{BO}{Bayesian Optimisation}
\acro{RL}{Reinforcement Learning}
\acro{ES}{Evolutionary Search}
\acro{MLP}{Multilayer Perceptron}
\acro{STE}{Straight-Through Estimator}
\acro{DARTS}{Differentiable Architecture Search}
\acro{SCP}{Secondary Carrier Prediction}
\acro{VoD}{Video on Demand}
\acro{GS}{Gumbel-Softmax}
\acro{STE}{Straight Through Estimator}
\acro{GS-STE}{Gumbel-Softmax with Straight Through Estimator}
\acro{RSRP}{Reference Signal Received Power}
\acro{RSS}{Received Signal Strength}
\acro{NWDAF}{Network Data Analytics Function}
\acro{RAN}{Radio Access Network}
\acro{DN}{Data Network}
\acro{LCM}{Life Cycle Management}
\acro{UE}{User Equipment}

\end{acronym}

\title{Automated Model Design using Gated Neuron Selection in Telecom}
\maketitle

\renewcommand{\thefootnote}{\IEEEauthorrefmark{1}}
\footnotetext{Work performed while at Ericsson Research.}
\renewcommand{\thefootnote}{\arabic{footnote}}

\renewcommand{\thefootnote}{}
\footnotetext{This work has been accepted for publication in IEEE/IFIP Network Operations and Management Symposium 2026. The final published version will be available via IEEE Xplore.}
\renewcommand{\thefootnote}{\arabic{footnote}}

\setcounter{footnote}{0}

\begin{abstract}

The telecommunications industry is experiencing rapid growth in adopting deep learning for critical tasks such as traffic prediction, signal strength prediction, and quality of service optimisation. However, designing neural network architectures for these applications remains challenging and time-consuming, particularly when targeting compact models suitable for resource-constrained network environments. Therefore, there is a need for automating the model design process to create high-performing models efficiently. This paper introduces \method~(Tabular Gated Neuron Selection), a novel gradient-based Neural Architecture Search (NAS) method specifically tailored for tabular data in telecommunications networks.
We evaluate \method~across multiple telecommunications and generic tabular datasets, demonstrating improvements in prediction performance while reducing the architecture size by 51--82\% and reducing the search time by up to 36x compared to state-of-the-art tabular NAS methods.
Integrating TabGNS into the model lifecycle management enables automated design of neural networks throughout the lifecycle, accelerating deployment of ML solutions in telecommunications networks.

\end{abstract}


\section{Introduction} 

The rapid evolution of telecommunication networks has accelerated the adoption of \ac{AI} and \ac{ML}, advancing toward the AI-native vision~\cite{iovene_detailed_2023}. Neural networks are increasingly leveraged for tasks such as traffic prediction~\cite{zhang_deep_2019}, received signal strength estimation~\cite{farooq_multi-task_2024}, and service performance prediction~\cite{yanggratoke_predicting_2015}. These models are expected to be deployed across diverse environments, ranging from the \ac{UE} to the \ac{RAN} and Core, each with distinct constraints on computational resources and requirements on model performance and inference latency.
Designing neural network architectures that meet these requirements is a complex and time-consuming process, often requiring significant manual effort and domain expertise. It is further pronounced by the need for continuous model maintenance and retraining as part of the model \ac{LCM} --- which should be fully automated to ensure sustainability over time.
Automating the model design is essential to achieve fully automated model \ac{LCM}.

\begin{figure}[t]
    \centering
    \includegraphics[width=0.95\columnwidth]{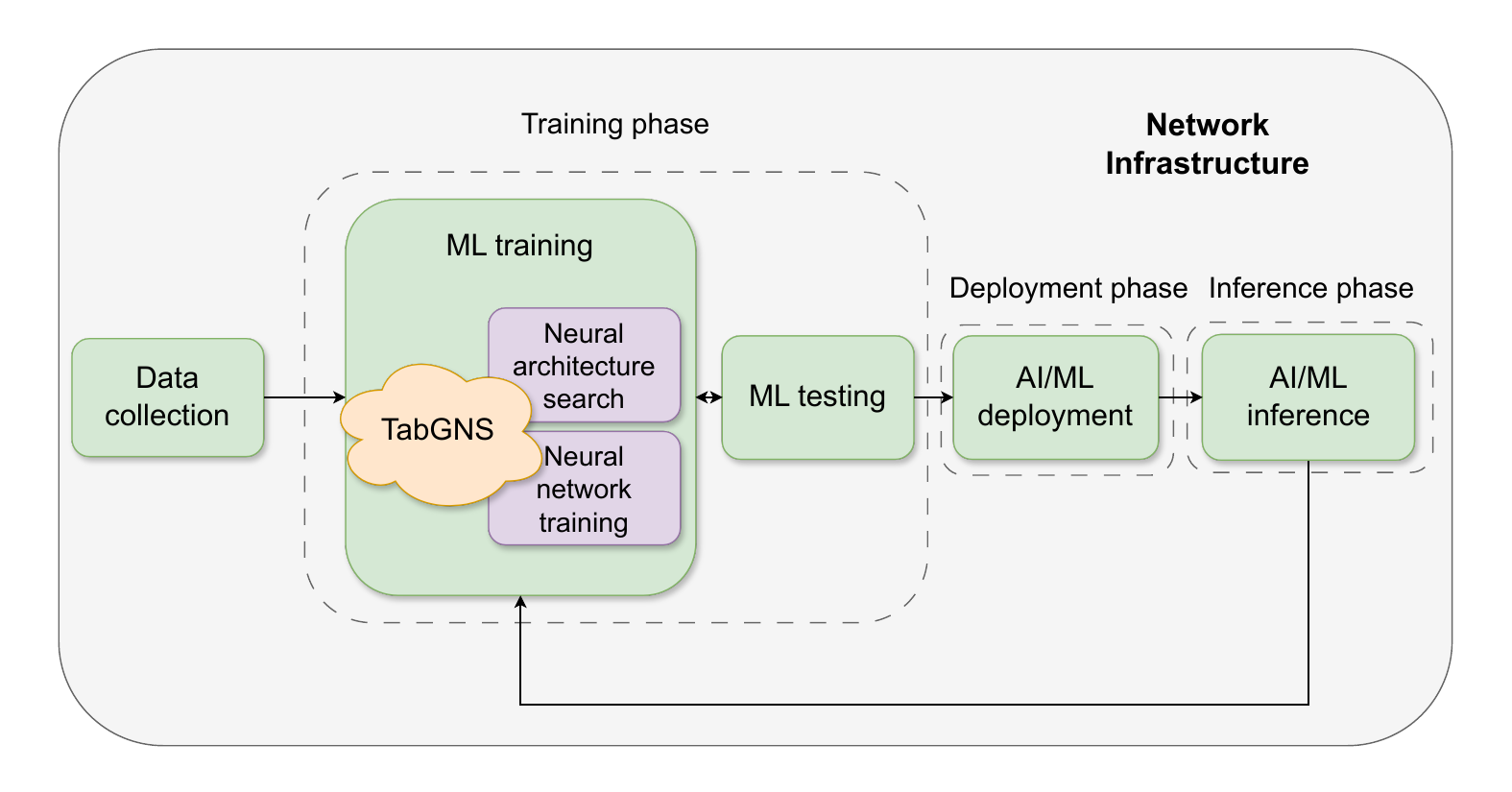}
    \caption{Workflow of the operational steps in the ML model lifecycle management~\cite{etsi_study_2024} augmented with TabGNS for automated neural architecture search and training.}
    \label{fig:front-diagram}
\end{figure}

\textit{\ac{NAS}}~\cite{zoph_neural_2017} is a promising approach for automating the discovery of high-performing neural network architectures. It employs learning algorithms to explore a predefined search space that selects a high-performing architecture while respecting constraints such as, inference latency, and model size~\cite{orucu_multi-objective_2024}. NAS has demonstrated success in domains such as computer vision and natural language processing. 
However, existing research for telecom has mainly focused on non-tabular data~\cite{wang_evolutionary_2021, zhang_nas-amr_2022}, despite the abundance of tabular data in telecom systems~\cite{chui_notes_2018}.

In this paper, we introduce \textit{Tabular Gated Neuron Selection (TabGNS)}\footnote{\texttt{https://github.com/EricssonResearch/tabgns}}, a novel gradient-based NAS method tailored for tabular data in telecommunications.
TabGNS can seamlessly be integrated into existing ML model \ac{LCM} frameworks in 3GPP or O-RAN based networks. An illustrative example is provided in Fig.~\ref{fig:front-diagram}. During the training phase, TabGNS performs joint architecture search and model training. The resulting model is deployed and used for inference. Continuous monitoring of model performance can then trigger retraining, for example due to changes in data distribution or resource availability, enabling re-optimization over both architecture and data.

To the best of our knowledge, TabGNS is the first NAS approach designed specifically for tabular telecom data and the first to apply gradient-based architecture search to \acp{MLP}, through a neuron-level gating mechanism.
In this paper, we have chosen to focus on \acp{MLP} as they remain widely used for tabular data in production telecom systems due to their efficiency, and proven performance. This choice ensures a uniform and controlled search space for fair comparison across all NAS methods, and simplifies the search space, making it easier to interpret and analyse the NAS algorithm's behaviour. We note that our gating mechanism is architecture-agnostic and can be extended to other layer types in future work.

\begin{figure}
    \centering
    \includegraphics[width=0.95\columnwidth]{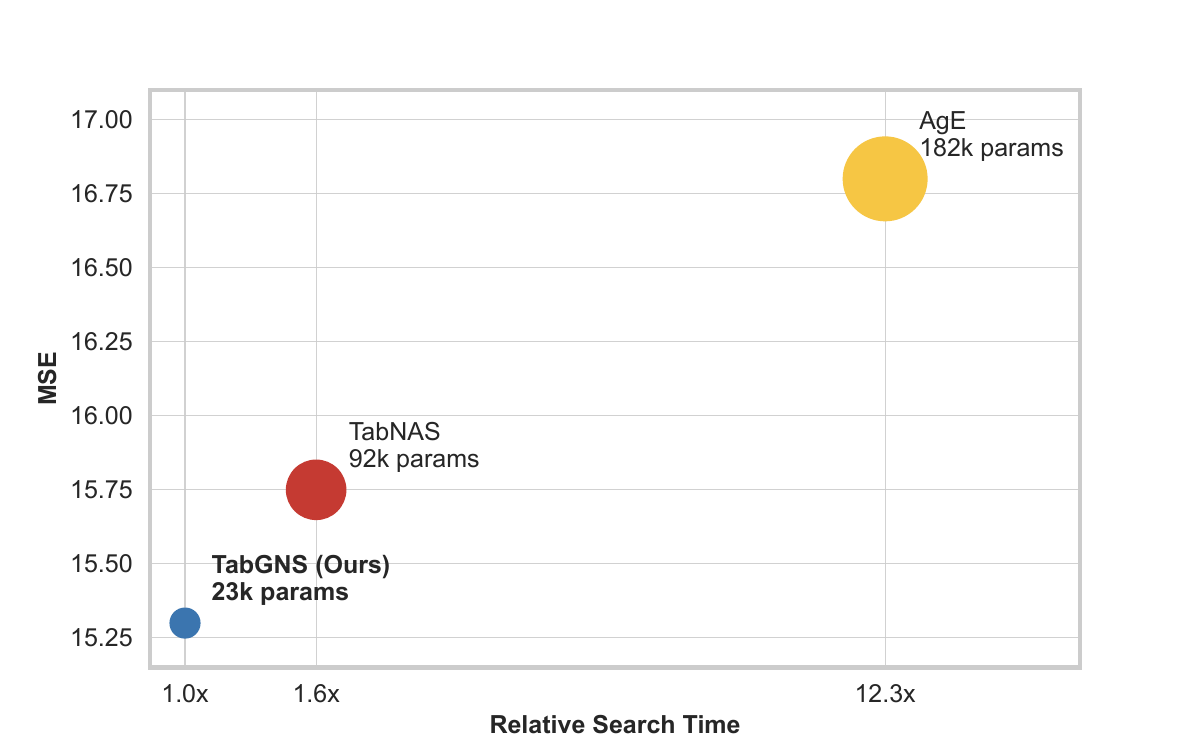}
    \caption{\method~outperforms previous tabular \ac{NAS} methods on all of the critical dimensions; prediction error, architecture size, and search time. Scatter-point size represents architecture size. Results for VoD dataset.}
    \label{fig:scatter-vod}
\end{figure}

At its core, TabGNS employs a progressive growth strategy, starting from small architectures and gradually widening them using the learning algorithm to efficiently discover compact models with high predictive performance. More specifically, TabGNS improves the architectural design of MLPs for tabular data along three key dimensions: (1) fast and efficient architecture discovery, (2) resulting compact model size, and (3) high predictive performance. Fig.~\ref{fig:scatter-vod} provides an indicative example (search time, mean square error, and model size) of the superiority of TabGNS compared to other state-of-the-art approaches.


Our contributions are as follows:
\begin{itemize}
  \item Identification of the need for automated MLP design in telecom networks, with particular interest in tabular data.
  \item Introduction of \method, a method for gradient-based neural architecture search through a neuron-level gating mechanism. 
  \item Comprehensive evaluation on multiple datasets demonstrating significant improvements compared to state-of-the-art --- matching or surpassing prediction performance while reducing model size by $51-82\%$, and accelerating search time by up to 36 times.
\end{itemize}

The remainder of this paper is organized as follows: Section~\ref{sec:related} reviews related work in \ac{NAS} and its applications in tabular and telecom data. Section~\ref{sec:problem} formalizes the problem description. Section~\ref{sec:method} presents our proposed method, \method, in detail. Sections~\ref{sec:data} and~\ref{sec:eval} describe our datasets, experimental setup, and evaluation methodology, and present our results and comparative analysis. Finally, Section~\ref{sec:conclusion} concludes the paper and discusses future research directions.

\begin{figure*}[ht]
  \centering

  \definecolor{oncolor}{RGB}{96, 104, 250}
  \definecolor{offcolor}{RGB}{250, 104, 96}
  \definecolor{neuroncolor}{RGB}{246, 198, 68}

  \colorlet{neuronFillColor}{neuroncolor}
  \colorlet{neuronBorderColor}{neuroncolor!75!black!90}
  \colorlet{connectionColor}{black}
  \colorlet{inactiveNeuronColor}{white}

  \colorlet{color20}{oncolor!20!offcolor} 
  \colorlet{color40}{oncolor!40!offcolor} 
  \colorlet{color60}{oncolor!60!offcolor} 
  \colorlet{color80}{oncolor!80!offcolor} 

  \def\initialcolors{"color20","color20","color20","color20","color20","color20","color20","color20","color20","color20","color20","color20"}
  \def\middlecolors{"color40","color60","color60","offcolor","color80","color80","color20","color60","color80","color80","color60","color20"}
  \def\finalcolors{"offcolor","oncolor","offcolor","oncolor","oncolor","offcolor","oncolor","offcolor","oncolor","oncolor","offcolor","offcolor"}

  \def\baseThicknesses{"0.1","0.1","0.1","0.1"}

  \def\notused{"black","black","black","black","black","black","black","black","black","black","black","black"}

  \newcommand{\NeuralNetwork}[8]{
    \def\xoffset{#1}
    \def\yoffset{#2}
    \def\scale{#3}
    \def\boxColors{#4}
    \def\connThicknesses{#5}
    \def\varyWeights{#6}
    \def\patternID{#7}
    \def\showBoxes{#8}
    
    \pgfmathsetseed{\patternID}
    
    \begin{scope}[
      shift={(\xoffset,\yoffset)},
      scale=\scale,
      neuron/.style={circle, draw=neuronBorderColor, fill=neuronFillColor, inner sep=0pt, minimum size=0.6cm*\scale},
      inactive/.style={circle, draw=neuronBorderColor, fill=inactiveNeuronColor, inner sep=0pt, minimum size=0.6cm*\scale},
      box/.style={rectangle, draw, minimum size=1.1cm*\scale},
      colorbox/.style={box, #4!70!black, fill=#4!30},
      conn/.style={connectionColor}
    ]
    
    \coordinate (i1) at (0,0.25);
    \coordinate (i2) at (0,-1.25);
    \coordinate (i3) at (0,-2.75);
    
    \foreach \i in {1,2,3,4} {
      \coordinate (h1\i) at (2,1.5*\i-5);
      
      \coordinate (h2\i) at (4,1.5*\i-5);
      
      \coordinate (h3\i) at (6,1.5*\i-5);
    }
    
    \coordinate (o1) at (8,-0.5);
    \coordinate (o2) at (8,-2);

    \foreach \i in {1,2,3,4} {
      \pgfmathsetmacro{\thicknessindex}{mod(\i-1,4)+1}
      \pgfmathparse{{\connThicknesses}[\thicknessindex-1]}
      \edef\baseThick{\pgfmathresult}
      
      \ifnum\i=1
        \foreach \j in {1,2,3}
          \foreach \k in {1,2,3,4} {
            \ifnum\varyWeights=1
              \pgfmathsetmacro{\randFactor}{0.1+rand}
              \draw[connectionColor, line width=0.5pt*\randFactor+0.5pt] (i\j) -- (h1\k);
            \else
              \draw[connectionColor, line width=0.4pt] (i\j) -- (h1\k);
            \fi
          }
      \else\ifnum\i=2
        \foreach \j in {1,2,3,4}
          \foreach \k in {1,2,3,4} {
            \ifnum\varyWeights=1
              \pgfmathsetmacro{\randFactor}{0.1+rand}
              \draw[connectionColor, line width=0.5pt*\randFactor+0.5pt] (h1\j) -- (h2\k);
            \else
              \draw[connectionColor, line width=0.4pt] (h1\j) -- (h2\k);
            \fi
          }
      \else\ifnum\i=3
        \foreach \j in {1,2,3,4}
          \foreach \k in {1,2,3,4} {
            \ifnum\varyWeights=1
              \pgfmathsetmacro{\randFactor}{0.1+rand}
              \draw[connectionColor, line width=0.5pt*\randFactor+0.5pt] (h2\j) -- (h3\k);
            \else
              \draw[connectionColor, line width=0.4pt] (h2\j) -- (h3\k);
            \fi
          }
      \else\ifnum\i=4
        \foreach \j in {1,2,3,4}
          \foreach \k in {1,2} {
            \ifnum\varyWeights=1
              \pgfmathsetmacro{\randFactor}{0.1+rand}
              \draw[connectionColor, line width=0.5pt*\randFactor+0.5pt] (h3\j) -- (o\k);
            \else
              \draw[connectionColor, line width=0.4pt] (h3\j) -- (o\k);
            \fi
          }
      \fi\fi\fi\fi
    }
    
    \ifnum\showBoxes=1
      \foreach \i in {1,2,3,4} {
        \pgfmathsetmacro{\colorindex}{mod(\i-1,12)+1}
        \pgfmathparse{{\boxColors}[\colorindex-1]}
        \edef\boxcolor{\pgfmathresult}
        
        \node[box, \boxcolor!70!black, fill=\boxcolor] (hb1\i) at (2,1.5*\i-5) {};
        
        \pgfmathsetmacro{\colorindex}{mod(\i+3,12)+1}
        \pgfmathparse{{\boxColors}[\colorindex-1]}
        \edef\boxcolor{\pgfmathresult}
        \node[box, \boxcolor!70!black, fill=\boxcolor] (hb2\i) at (4,1.5*\i-5) {};
        
        \pgfmathsetmacro{\colorindex}{mod(\i+7,12)+1}
        \pgfmathparse{{\boxColors}[\colorindex-1]}
        \edef\boxcolor{\pgfmathresult}
        \node[box, \boxcolor!70!black, fill=\boxcolor] (hb3\i) at (6,1.5*\i-5) {};
      }
    \fi
    
    \node[neuron] at (0,0.25) {};
    \node[neuron] at (0,-1.25) {};
    \node[neuron] at (0,-2.75) {};
    
    \foreach \i in {1,2,3,4} {
      \node[neuron] at (2,1.5*\i-5) {};
      
      \node[neuron] at (4,1.5*\i-5) {};
      
      \node[neuron] at (6,1.5*\i-5) {};
    }
    
    \node[neuron] at (8,-0.5) {};
    \node[neuron] at (8,-2) {};
    
    \end{scope}
  }

  \newcommand{\ConditionalNeuralNetwork}[7]{
    \def\xoffset{#1}
    \def\yoffset{#2}
    \def\scale{#3}
    \def\boxColors{#4}
    \def\connThicknesses{#5}
    \def\varyWeights{#6}
    \def\patternID{#7}
    
    \pgfmathsetseed{\patternID}
    
    \begin{scope}[
      shift={(\xoffset,\yoffset)},
      scale=\scale,
      neuron/.style={circle, draw=neuronBorderColor, fill=neuronFillColor, inner sep=0pt, minimum size=0.6cm*\scale},
      inactive/.style={circle, draw=neuronBorderColor, fill=inactiveNeuronColor, inner sep=0pt, minimum size=0.6cm*\scale},
      box/.style={rectangle, draw, minimum size=1.1cm*\scale},
      colorbox/.style={box, #4!70!black, fill=#4},
      conn/.style={connectionColor}
    ]
    
    \coordinate (i1) at (0,0.25);
    \coordinate (i2) at (0,-1.25);
    \coordinate (i3) at (0,-2.75);
    
    \foreach \i in {1,2,3,4} {
      \coordinate (h1\i) at (2,1.5*\i-5);
      
      \coordinate (h2\i) at (4,1.5*\i-5);
      
      \coordinate (h3\i) at (6,1.5*\i-5);
    }
    
    \coordinate (o1) at (8,-0.5);
    \coordinate (o2) at (8,-2);

    \def\h1active{0,1,0,1}
    \def\h2active{1,0,1,0}
    \def\h3active{0,1,0,1}

    \foreach \j in {1,2,3} {
      \foreach \k in {1,2,3,4} {
        \ifnum\k=1
          \def\isActive{0}
        \else\ifnum\k=2
          \def\isActive{1}
        \else\ifnum\k=3
          \def\isActive{0}
        \else
          \def\isActive{1}
        \fi\fi\fi
        
        \ifnum\isActive=1
          \ifnum\varyWeights=1
            \pgfmathsetmacro{\randFactor}{0.1+rand}
            \draw[connectionColor, line width=0.5pt*\randFactor+0.5pt] (i\j) -- (h1\k);
          \else
            \draw[connectionColor, line width=0.4pt] (i\j) -- (h1\k);
          \fi
        \fi
      }
    }
    
    \foreach \j in {1,2,3,4} {
      \ifnum\j=1
        \def\sourceActive{0}
      \else\ifnum\j=2
        \def\sourceActive{1}
      \else\ifnum\j=3
        \def\sourceActive{0}
      \else
        \def\sourceActive{1}
      \fi\fi\fi
      
      \foreach \k in {1,2,3,4} {
        \ifnum\k=1
          \def\targetActive{1}
        \else\ifnum\k=2
          \def\targetActive{0}
        \else\ifnum\k=3
          \def\targetActive{1}
        \else
          \def\targetActive{0}
        \fi\fi\fi
        
        \ifnum\sourceActive=1
          \ifnum\targetActive=1
            \ifnum\varyWeights=1
              \pgfmathsetmacro{\randFactor}{0.1+rand}
              \draw[connectionColor, line width=0.5pt*\randFactor+0.5pt] (h1\j) -- (h2\k);
            \else
              \draw[connectionColor, line width=0.4pt] (h1\j) -- (h2\k);
            \fi
          \fi
        \fi
      }
    }
    
    \foreach \j in {1,2,3,4} {
      \ifnum\j=1
        \def\sourceActive{1}
      \else\ifnum\j=2
        \def\sourceActive{0}
      \else\ifnum\j=3
        \def\sourceActive{1}
      \else
        \def\sourceActive{0}
      \fi\fi\fi
      
      \foreach \k in {1,2,3,4} {
        \ifnum\k=1
          \def\targetActive{1}
        \else\ifnum\k=2
          \def\targetActive{1}
        \else\ifnum\k=3
          \def\targetActive{0}
        \else
          \def\targetActive{0}
        \fi\fi\fi
        
        \ifnum\sourceActive=1
          \ifnum\targetActive=1
            \ifnum\varyWeights=1
              \pgfmathsetmacro{\randFactor}{0.1+rand}
              \draw[connectionColor, line width=0.5pt*\randFactor+0.5pt] (h2\j) -- (h3\k);
            \else
              \draw[connectionColor, line width=0.4pt] (h2\j) -- (h3\k);
            \fi
          \fi
        \fi
      }
    }
    
    \foreach \j in {1,2,3,4} {
      \ifnum\j=1
        \def\sourceActive{1}
      \else\ifnum\j=2
        \def\sourceActive{1}
      \else\ifnum\j=3
        \def\sourceActive{0}
      \else
        \def\sourceActive{0}
      \fi\fi\fi
      
      \ifnum\sourceActive=1
        \foreach \k in {1,2} {
          \ifnum\varyWeights=1
            \pgfmathsetmacro{\randFactor}{0.1+rand}
            \draw[connectionColor, line width=0.5pt*\randFactor+0.5pt] (h3\j) -- (o\k);
          \else
            \draw[connectionColor, line width=0.4pt] (h3\j) -- (o\k);
          \fi
        }
      \fi
    }
    
    \node[neuron] at (0,0.25) {};
    \node[neuron] at (0,-1.25) {};
    \node[neuron] at (0,-2.75) {};
    
    \node[inactive] at (2,1.5*1-5) {};
    \node[neuron] at (2,1.5*2-5) {};
    \node[inactive] at (2,1.5*3-5) {};
    \node[neuron] at (2,1.5*4-5) {};
    
    \node[neuron] at (4,1.5*1-5) {};
    \node[inactive] at (4,1.5*2-5) {};
    \node[neuron] at (4,1.5*3-5) {};
    \node[inactive] at (4,1.5*4-5) {};
    
    \node[neuron] at (6,1.5*1-5) {};
    \node[neuron] at (6,1.5*2-5) {};
    \node[inactive] at (6,1.5*3-5) {};
    \node[inactive] at (6,1.5*4-5) {};
    
    \node[neuron] at (8,-0.5) {};
    \node[neuron] at (8,-2) {};
    
    \end{scope}
  }

  \newcommand{\RegularNeuralNetwork}[7]{
    \NeuralNetwork{#1}{#2}{#3}{#4}{#5}{#6}{#7}{0}
  }

  \newcommand{\EmptyBox}[3]{
    \def\xoffset{#1}
    \def\yoffset{#2}
    \def\scale{#3}
    
    \begin{scope}[shift={(\xoffset,\yoffset)}, scale=\scale]
      \draw[dashed, thick, rounded corners] (0,-4) rectangle (8,1);
    \end{scope}
  }

  \newcommand{\TwoRowNetworks}[5][]{
    
    \def\colorSchemeIndices{#2}
    \def\showBoxesParam{#3}
    \def\patternThree{#4}
    \def\patternFour{#5}
    
    \pgfkeyssetvalue{/twoRows/width}{\textwidth}
    \pgfqkeys{/twoRows}{#1}
    \pgfkeysgetvalue{/twoRows/width}{\figureWidth}
    
    \pgfmathsetmacro{\networkWidth}{9}
    \pgfmathsetmacro{\totalUnitsTopRow}{4*\networkWidth + 3*2.5}
    \pgfmathsetmacro{\overallScale}{\figureWidth/\totalUnitsTopRow/28.453*0.85} 
    
    \pgfmathsetmacro{\networkSpacing}{1.5*\overallScale}
    
    \pgfmathsetmacro{\netOneX}{0}
    \pgfmathsetmacro{\netTwoX}{\netOneX + \networkWidth*\overallScale + \networkSpacing}
    \pgfmathsetmacro{\emptyBoxX}{\netTwoX + \networkWidth*\overallScale + \networkSpacing}
    \pgfmathsetmacro{\netSixX}{\emptyBoxX + \networkWidth*\overallScale + \networkSpacing}
    \pgfmathsetmacro{\netSevenX}{\netSixX + \networkWidth*\overallScale + \networkSpacing}
    
    \pgfmathsetmacro{\bottomRowWidth}{4*\networkWidth*\overallScale + 3*\networkSpacing} 
    \pgfmathsetmacro{\bottomRowStart}{(\totalUnitsTopRow*\overallScale - \bottomRowWidth)/2}
    \pgfmathsetmacro{\netThreeX}{\bottomRowStart}
    \pgfmathsetmacro{\netFourX}{\netThreeX + \networkWidth*\overallScale + \networkSpacing}
    \pgfmathsetmacro{\netFiveX}{\netFourX + \networkWidth*\overallScale + \networkSpacing}
    \pgfmathsetmacro{\netEightX}{\netFiveX + \networkWidth*\overallScale + \networkSpacing} 
    
    \pgfmathsetmacro{\secondRowY}{-10*\overallScale}
    
    \pgfmathparse{{\colorSchemeIndices}[0]}
    \edef\colorIndexOne{\pgfmathresult}
    \pgfmathparse{{\colorSchemeIndices}[1]}
    \edef\colorIndexTwo{\pgfmathresult}
    \pgfmathparse{{\colorSchemeIndices}[2]}
    \edef\colorIndexThree{\pgfmathresult}
    \pgfmathparse{{\colorSchemeIndices}[3]}
    \edef\colorIndexFour{\pgfmathresult}
    
    \ifnum\colorIndexOne=1\def\schemeOne{\initialcolors}\else\ifnum\colorIndexOne=2\def\schemeOne{\middlecolors}\else\ifnum\colorIndexOne=3\def\schemeOne{\finalcolors}\else\def\schemeOne{\notused}\fi\fi\fi
    
    \ifnum\colorIndexTwo=1\def\schemeTwo{\initialcolors}\else\ifnum\colorIndexTwo=2\def\schemeTwo{\middlecolors}\else\ifnum\colorIndexTwo=3\def\schemeTwo{\finalcolors}\else\def\schemeTwo{\notused}\fi\fi\fi
    
    \ifnum\colorIndexThree=1\def\schemeThree{\initialcolors}\else\ifnum\colorIndexThree=2\def\schemeThree{\middlecolors}\else\ifnum\colorIndexThree=3\def\schemeThree{\finalcolors}\else\def\schemeThree{\notused}\fi\fi\fi
    
    \ifnum\colorIndexFour=1\def\schemeFour{\initialcolors}\else\ifnum\colorIndexFour=2\def\schemeFour{\middlecolors}\else\ifnum\colorIndexFour=3\def\schemeFour{\finalcolors}\else\def\schemeFour{\notused}\fi\fi\fi
    
    \RegularNeuralNetwork{\netOneX}{0}{\overallScale}{\initialcolors}{\baseThicknesses}{0}{0}
    
    \draw[->,thick] 
      (\netOneX + \networkWidth*\overallScale , -0.5) -- 
      (\netTwoX -0.2, -0.5);
      
    \NeuralNetwork{\netTwoX}{0}{\overallScale}{\initialcolors}{\baseThicknesses}{0}{0}{1}
    
    \draw[->,thick] 
      (\netTwoX + \networkWidth*\overallScale , -0.5) -- 
      (\emptyBoxX -0.2, -0.5);
    
    \EmptyBox{\emptyBoxX}{0}{\overallScale}
    \node[text width=200*\overallScale, align=center] at (\emptyBoxX + 4*\overallScale, -0.5) 
      {Architecture search and weight training};
      
    \coordinate (emptyBoxBottomLeft) at (\emptyBoxX, -4*\overallScale);
    \coordinate (emptyBoxBottomRight) at (\emptyBoxX + 8*\overallScale, -4*\overallScale);
    
    \draw[->,thick] 
      (\emptyBoxX + \networkWidth*\overallScale , -0.5) -- 
      (\netSixX -0.2, -0.5);
    
    \NeuralNetwork{\netSixX}{0}{\overallScale}{\schemeThree}{\baseThicknesses}{1}{\patternThree}{\showBoxesParam}
    
    \draw[->,thick] 
      (\netSixX + \networkWidth*\overallScale , -0.5) -- 
      (\netSevenX -0.2, -0.5);
      
    \ConditionalNeuralNetwork{\netSevenX}{0}{\overallScale}{\schemeFour}{\baseThicknesses}{1}{\patternFour}
    
    \pgfmathsetmacro{\boxLeft}{\netThreeX - \networkSpacing}
    \pgfmathsetmacro{\boxRight}{\netEightX + \networkWidth*\overallScale + \networkSpacing}
    \pgfmathsetmacro{\boxTop}{\secondRowY + 2.5*\overallScale} 
    \pgfmathsetmacro{\boxBottom}{\secondRowY - 5*\overallScale} 
    
    \coordinate (bottomBoxTopLeft) at (\boxLeft, \boxTop);
    \coordinate (bottomBoxTopRight) at (\boxRight, \boxTop);
    
    \draw[dashed, thick, rounded corners] (\boxLeft,\boxTop) rectangle (\boxRight,\boxBottom);
    
    \NeuralNetwork{\netThreeX}{\secondRowY}{\overallScale}{\schemeOne}{\baseThicknesses}{0}{0}{\showBoxesParam}

    \draw[->,thick] 
      (\netThreeX + \networkWidth*\overallScale , \secondRowY -0.5) -- 
      (\netFourX -0.2, \secondRowY -0.5);

    \NeuralNetwork{\netFourX}{\secondRowY}{\overallScale}{\schemeTwo}{\baseThicknesses}{1}{1}{\showBoxesParam}

    \draw[->,thick] 
      (\netFourX + \networkWidth*\overallScale , \secondRowY -0.5) -- 
      (\netFiveX -0.2, \secondRowY -0.5);
    
    \node[font=\LARGE] at (\netFiveX + 4*\overallScale, \secondRowY - 0.5) {\ldots};

    \draw[->,thick] 
      (\netFiveX + \networkWidth*\overallScale , \secondRowY -0.5) -- 
      (\netEightX -0.2, \secondRowY -0.5);
    
    \NeuralNetwork{\netEightX}{\secondRowY}{\overallScale}{\schemeThree}{\baseThicknesses}{1}{\patternThree}{\showBoxesParam}
    
    \draw[thin] (emptyBoxBottomLeft) -- (bottomBoxTopLeft);
    
    \draw[thin] (emptyBoxBottomRight) -- (bottomBoxTopRight);
  }

  \newcommand{\ColorSpectrumLegend}[5]{
    \def\legX{#1}
    \def\legY{#2}
    \def\legWidth{#3}
    \def\legHeight{#4}
    \def\numSegments{#5}
    
    \pgfmathsetmacro{\segHeight}{\legHeight/\numSegments}

    \colorlet{topc}{oncolor}
    \colorlet{botc}{offcolor}
    
    \foreach \i in {0,...,\numSegments} {
      \pgfmathsetmacro{\ypos}{\legY + \i*\segHeight}
      \pgfmathsetmacro{\nextYpos}{\legY + (\i+1)*\segHeight}
      \pgfmathsetmacro{\blendFactor}{100*\i/\numSegments}
      \fill[topc!\blendFactor!botc] (\legX, \ypos) rectangle (\legX+\legWidth, \nextYpos);
    }
    
    \node[anchor=center, font=\small] at (\legX + \legendWidth, \legY+\legHeight+0.4) {$p_{\text{open}}=1.0$};
    \node[anchor=center, font=\small] at (\legX + \legendWidth, \legY-0.2) {$p_{\text{open}}=0.0$};

  }

  \begin{tikzpicture}
    \TwoRowNetworks[width=0.95\textwidth]{2,2,3,3}{1}{5}{1}
    
    \pgfmathsetmacro{\legendHeight}{\boxTop - \boxBottom -0.6}
    \pgfmathsetmacro{\legendWidth}{0.6}
    \pgfmathsetmacro{\legendX}{\boxRight + 1}
    \pgfmathsetmacro{\legendY}{\boxBottom + 0.2}
    
    \ColorSpectrumLegend{\legendX}{\legendY}{\legendWidth}{\legendHeight}{10}

    \node[anchor=north west, font=\small\itshape] at (\netTwoX - 0.1, 0.9) {a)};
    \node[anchor=north west, font=\small\itshape] at (\boxLeft + 0.2, \boxTop - 0.1) {b)};
    \node[anchor=north west, font=\small\itshape] at (\netSixX - 0.1, 0.9) {c)};
    \node[anchor=north west, font=\small\itshape] at (\netSevenX - 0.1, 0.9) {d)};
  \end{tikzpicture}

\caption{Representation of the learning process of \method. (a) Gate for each neuron is initialised to a be open at a low probability. (b) Architecture and weights are iteratively updated until convergence. (c) Neurons with gates with probability $p_{\text{open}}>0.5$ are extracted together with their weights. (d) Final trained fully connected neural network, can be further fine tuned if needed.}
\label{fig:diagram}
\end{figure*}
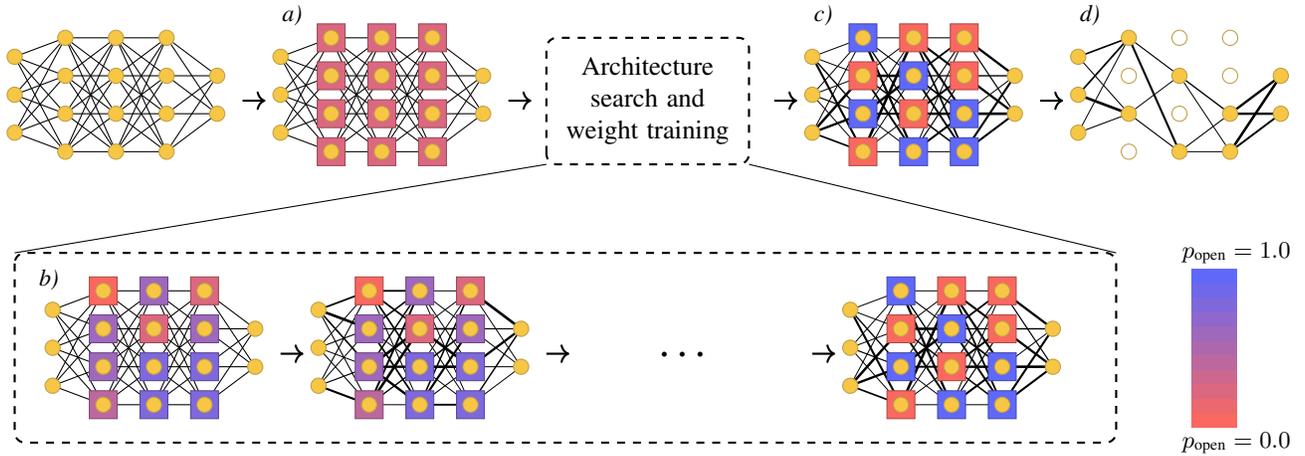

\section{Related Work} 
\label{sec:related}


Finding deep learning architectures for tabular data that can outperform tree-based ensemble methods has been an active research area and several deep learning architectures such as transformer based architectures~\cite{gorishniy_revisiting_2023, arik_tabnet_2020} have been developed.  Although these architectures show promising results, they do not always outperform vanilla MLP models while requiring significantly higher computational resources~\cite{gorishniy_tabm_2024}. Recently MLP based ensemble architectures have shown competitive results to tree based methods while being more efficient than more advanced and complicated architectures~\cite{gorishniy_tabm_2024, rubachev_tabred_2024}. In this paper, we focus on automatically finding the best MLP architecture for tabular telecom data using NAS. As a future work one could enhance the selected MLP with ensemble methods to further improve the model performance.

NAS has achieved remarkable success in finding models with state-of-the-art accuracy across various tasks, particularly for image and text data \cite{elsken_neural_2019}. Using NAS in telecom has also received some attention. Wang et al.~\cite{wang_evolutionary_2021} use NAS to find the best convolutional neural network for traffic classification, and NAS-AMR~\cite{zhang_nas-amr_2022} is proposed for automatic modulation recognition using simulated signal datasets. Cooperatively optimising data collection and NAS for IoT devices and image data was studied by Yin et al.~\cite{yin_dynamic_2022}. While these papers explored NAS for telecom use cases, they all used unstructured datasets such as images and signals. However, the majority of the \ac{ML} tasks for network management are based on tabular datasets~\cite{chui_notes_2018}, which is our focus in this paper.

According to Yang et al.~\cite{yang_tabnas_2022}, NAS for tabular data has received limited attention due to a lack of understanding of promising architectures and a lack of relevant datasets.
Although few, there is some work in the area, including AgEBO-tabular~\cite{egele_agebo-tabular_2021} and  TabNAS~\cite{yang_tabnas_2022}, both of which have been used as baselines in this paper. Similar to AgEBO-tabular, Xing et al.~\cite{xing_anytime_2024} use Aging Evolution NAS for tabular data and \cite{das_fairer_2023} is designed with focus on fairness; therefore are out of scope for this paper. We utilise the \ac{NAS} part of AgEBO-tabular and not \ac{BO} for hyperparameter tuning therefore we refer to it as AgE.


DARTS~\cite{liu_darts_2019}, being a gradient descent based method, has inspired numerous extensions that leverage continuous search spaces for faster search times. Several works use Gumbel-Softmax (GS)~\cite{jang_categorical_2017}, or Straight-Through Estimator (STE)~\cite{bengio_estimating_2013} functions to enable gradient-based optimisation: ProxylessNAS~\cite{cai_proxylessnas_2019}, SNAS~\cite{xie_snas_2019}, Dong et al.~\cite{dong_searching_2019}, and FBNet~\cite{wu_fbnet_2019} use GS to select between different layer types, while GS-NAS~\cite{pang_gumbel-softmax_2022} applies it to select layer widths for Deep Belief Networks. However, these DARTS-based methods have primarily focused on computer vision and natural language processing tasks. Our work contributes to the state of the art by being the first gradient-based NAS method specifically designed for tabular data problems.

Other distantly related works are Compete to compute~\cite{srivastava_compete_2013} which turns off neurons given its performance relative to its neighbour and Dropout~\cite{srivastava_dropout_2014} which randomly drops neurons during training. Neither of these works are applied to architecture search. Pruning techniques~\cite{cheng_survey_2024} also aim to minimise the architectures parameter count while maintaining good performance. Specifically, Pruning-as-Search~\cite{li_pruning-as-search_2022} aims to find a pruned architecture using \ac{NAS}. However, these type of methods start from a trained model and prune weights instead of neurons. Our method, on the other hand, trains the models while selecting the architecture and keeps the final model fully-connected which is a structure more suitable for GPUs.


\section{Problem Description} 
\label{sec:problem}

Deploying effective neural networks in telecom systems requires careful architecture selection to balance competing requirements such as limited computational resources and high predictive performance --- a task traditionally performed manually. To address this time-consuming and inefficient process, there is a need for an automated architecture selection process that can efficiently determine a high-performing neural network structure without human intervention. This process should optimise \ac{ML} models as part of fully automated \ac{LCM}, to achieve high accuracy while minimising computational overhead both during architecture search and at model inference.

Given an \ac{ML} task, we define a search space $S$ parametrised by the number of hidden layers $L$ and the maximum width $W$ of fully connected layers. $S$ corresponds to the largest possible \ac{MLP} that contains $W^L$ unique smaller architectures. The objective is to find an architecture $A \subseteq S$ that is significantly smaller than $S$ and that achieves comparable or superior predictive performance to $S$ when both are trained. Thus, the optimisation objective is,

\begin{equation}
  \begin{array}{cl}
  \min _{A \subseteq S} & \mathcal{L}_{\text {val }}\left(w^*(A), A\right) \\
  \text { s.t. } & w^*(A)=\operatorname{argmin}_w \mathcal{L}_{\text {train }}(w, A) \\
  \end{array}
  \label{eq:darts-opt}
\end{equation}

where $w$ are the weights of the model, $\mathcal{L}$ is the loss of the model that is calculated on the training or validation data. The optimization process aims to select a subset of neurons within $S$ to construct $A$, reducing computational complexity while maintaining or enhancing model performance.
In this paper we introduce a novel search algorithm that can efficiently find a small architecture $A$ that is optimised for Eq.~\eqref{eq:darts-opt}.

\section{Method}
\label{sec:method}
This section provides the necessary background and presents Tabular Gated Neuron Selection (\method) --- a NAS method that automates parts of the model lifecycle management.

\subsection{Background}
\ac{NAS} involves searching for an optimal architecture by selecting from choices of layers and how they are to be connected. Since these choices are discrete, traditional gradient-based methods cannot directly handle them. There are several approaches that can enable the use of gradient-based methods on discrete variables. In this section, we introduce the methods used in this paper.

\subsubsection*{Gumbel-Softmax}

Gumbel-softmax~\cite{jang_categorical_2017} is a differentiable approximation of a categorical sampling. It allows backpropagation through random variables, which are otherwise discrete choices.
Each category's probability is represented by a logit which is summed with an independent random variable from a Gumbel distribution. The sum is divided by a temperature term $\tau$, which is used to control the sharpness of the distribution (the ``randomness'' in the samples). The softmax of this term is used to provide probabilities for each category. The process is defined by Eq.~\eqref{eq:gs},

\begin{equation} \label{eq:gs}
  y_i=\frac{\exp \left(\left(\log \left(\pi_i\right)+o_i\right) / \tau\right)}{\sum_{j=1}^k \exp \left(\left(\log \left(\pi_j\right)+o_j\right) / \tau\right)} \quad \text { for } i=1, \ldots, k
\end{equation}

where $\pi_i$ is the logit of the category $i$ out of $k$ categories and $o_i$ is a random sample from the Gumbel distribution.
As the temperature $\tau$ approaches zero, the distribution of $y$ approaches a categorical distribution.
Gumbel-softmax is frequently used together with the Straight-Through Estimator (STE).

\subsubsection*{Straight Through Estimator}
STE~\cite{bengio_estimating_2013} uses discrete variables such as argmax during the forward propagation, while in the backward propagation the gradient is approximated by assuming the operations are continuous. This approximation enables the use of standard backpropagation, despite the problem being discrete.

\subsubsection*{Gumbel Softmax with Straight Through Estimator}
In this variant of Gumbel-softmax, a discrete operation (one-hot encoding of arg-max of $y$) is used during the forward pass. During the backward pass, meanwhile, continuous values of $y$ (pre-encoding) are used.
Given,
\begin{equation}
  z_i = \text{one-hot}\left(\arg \max_j y_j\right)
\end{equation}

the gradient is approximated as,
\begin{equation}
  \frac{\partial L}{\partial \pi_i} \approx \frac{\partial L}{\partial z_i} \cdot \frac{\partial y_i}{\partial \pi_i}
\end{equation}

where $L$ is the loss, $\partial L / \partial z_i$ is gradient flowing into the hard output $z_i$ from subsequent layers and $\partial y_i / \partial \pi_i$ is the gradient of the Gumbel-softmax with respect to the logits. 

\subsubsection*{Differentiable Architecture Search}
DARTS~\cite{liu_darts_2019} is a well-established method for architecture search. It uses a continuous relaxation of the architecture search space to find a high-performing architecture.
The architecture search space is defined by various neural network operations, ranging from layer types (identity, convolution, linear) to layer-specific properties (kernel size, stride length, layer width).
While these are discrete options, DARTS makes them continuous by weighting them and propagating the weighted outputs of each operation. The weight of each option is characterised by a parameter which together are passed through a softmax function. These parameters are later updated using traditional gradient-based learning algorithms until they converge. Therefore, one of the multiple operations is chosen as the final operation. This process is done for every choice in the architecture search space.

In a DARTS learning process, the algorithm optimises the weights of the entire architecture, a \textit{SuperNet}~\cite{pham_efficient_2018} containing all possible architectures, on the training data-split and the architecture on the validation data-split. Storing all possible architectures in one SuperNet means that a portion of the weights between two different architectures are likely to overlap. This method of storing and training architectures, called \textit{weight sharing}, is frequently used due to its efficiency caused by not needing to retrain all the weights for every architecture being tested.
The training process iteratively alternates between weight optimisation and architecture optimisation, with each step performed on individual data batches.
Training continues until the architecture converges, with one operation for each selection of operations.

DARTS has proven effective for architectures such as ConvNets but faces some challenges when applied to fully connected networks for tabular data. Convolutional layers are agnostic to the input size thanks to the ``sliding-window'' behaviour of the convolution operation. This is not the case for fully connected layers, which require explicit definitions of the input and output dimensions.
Furthermore, this approach misses efficiency opportunities; for instance, when considering two candidate layers of widths 5 and 7, the weights in the width-5 layer represent a subset of those in the width-7 layer, yet conventional DARTS would train these independently without leveraging potential weight sharing.

\subsection{\method: Tabular Gated Neuron Selection}
We propose shifting from layer-level to neuron-level gradient-based architecture search. Our method begins with a fully connected SuperNet that includes all candidate neurons.
The final architecture is derived based on each neuron's contribution to predictions, effectively selecting a high-performing subset of neurons from the SuperNet.
Since different architectures are subsets of the SuperNet, many of the weights are shared and reused across the architectures. This overlap reduces the need to store and train separate sets of weights, improving training efficiency.
Moreover, this approach provides finer control over the search space, enabling more precise decisions than traditional layer-based approaches.

We implement the selection process by associating each neuron in the search space with a ``gate'' that combines a learnable parameter and an activation function. These gates control whether neurons are active or inactive. Although many activation functions can be used, we select Gumbel-softmax with Straight-Through Estimator (GS-STE) due to its widespread adoption and well-established theoretical understanding. Thus, \method~provides a learnable mechanism for sampling whether a neuron is included. During the training process, the probabilities of different gates change, reflecting the contribution of the corresponding neuron. At the end of the training process, the gates are binarised to determine which neurons are included in the final architecture.

Each gate requires only one parameter, $g^{(i,j)}$, representing the ``on'' state of the neuron; the ``off'' state is fixed to zero. The fixed value still participates in the stochastic process through the Gumbel distribution's randomness. Consequently, the GS-STE formulation, with many categories in Eq.~\eqref{eq:gs}, requires only two options. The activation probability of a single gate in position $(i, j)$ in the neural network can be expressed as;

\begin{equation} \label{eq:tabdarts-gate}
  p^{(i,j)} = \frac{\exp \left(\left(\log \left(g^{(i,j)}\right)+o_1\right) / \tau\right)}
  {\exp \left(\left(\log \left(g^{(i,j)}\right)+o_1\right) / \tau\right) + \exp {\left( o_2  / \tau \right)}}
\end{equation}

where $o_1$ and $o_2$ are the samples from the Gumbel distribution which are separately sampled for every gate.

The search space of \method, $S$, defines the largest possible neural network.
As visualised in Fig.~\ref{fig:diagram}, each hidden neuron is assigned a corresponding gate, where
each gate contains a parameter $g^{(i,j)}$ that determines the probability of activating its associated neuron. The complete set of these parameters, denoted as $\mathbf{g}$, defines our searchable architecture space.

\method~follows the standard DARTS training process of alternating optimization explained in background section. Weights of the model are updated using training data while gates remain frozen, then gates are updated using validation data while weights remain frozen. This iterative process ensures that architectural decisions (gates) and parameter learning (weights) are optimised independently.
Pseudo code for the training process is provided in Algorithms~\ref{algo:pseudocode} and \ref{algo:forward}. To explain this process more intuitively: During each forward pass, neurons are stochastically sampled based on their gate probabilities ($p_{open}$), and information flows only through the selected active neurons. The alternating optimization then updates either the weights or gates of these active neurons, depending on the current training phase.

\begin{algorithm}[t] 
\DontPrintSemicolon
\SetKwInOut{Input}{Input}
\SetKwInOut{Output}{Output}
\Input{Neural network $S$ with weights and biases $\mathbf{w}$, and training data $(X_{train}, Y_{train})$, validation data $(X_{valid}, Y_{valid})$}
\Output{Architecture A and its weights}

// Initialise gate parameters $\mathbf{g} = \{g^{(i,j)}\}$ for each neuron at position $(i,j)$\; 

\While{not converged}{
  // Create batches from training and validation data\;
  $\{B_{train}^i \} \leftarrow$ CreateBatches$(X_{train}, Y_{train})$\;
  $\{B_{valid}^i \} \leftarrow$ CreateBatches$(X_{valid}, Y_{valid})$\;
  
  \For{each batch $(B_{train}^i, B_{valid}^i)$}{
    // Step 1: Optimise network weights\;
    $\mathbf{g}.\text{trainable} \leftarrow \text{False}$\;
    $\mathbf{w}.\text{trainable} \leftarrow \text{True}$\;
    $\mathcal{L}_{train} \leftarrow$ ForwardPass$(\mathbf{w} \mid \mathbf{g}, B_{train}^i)$ \;
    Update weights $\mathbf{w} \leftarrow \mathbf{w} - \eta_w \nabla_{\mathbf{w}} \mathcal{L}_{train}$ \;
    
    // Step 2: Optimise architecture (gates)\;
    $\mathbf{g}.\text{trainable} \leftarrow \text{True}$\;
    $\mathbf{w}.\text{trainable} \leftarrow \text{False}$\;
    $\mathcal{L}_{valid} \leftarrow$ ForwardPass$(\mathbf{g} \mid \mathbf{w}, B_{valid}^i)$ \;
    Update gates $\mathbf{g} \leftarrow \mathbf{g} - \eta_g \nabla_{\mathbf{g}} \mathcal{L}_{valid}$ \;
  }
}

// Extract final architecture. $p$ calculated using Eq.~\eqref{eq:tabdarts-gate}\;
$A \leftarrow \{(i,j) \mid p_{open}^{(i,j)} \geq 0.5 \}$\;

\caption{{\sc \method}}
\label{algo:pseudocode}
\end{algorithm}

\begin{algorithm}[t] 
\DontPrintSemicolon
\SetKwInOut{Input}{Input}
\SetKwInOut{Output}{Output}
\Input{Weights $\mathbf{w}$, biases $b$, gate parameters $\mathbf{g}$, input data $X$, target data $Y$ and loss function \text{loss\_fn}}
\Output{Loss value $\mathcal{L}$}

// Initialise with input data\;
$h \leftarrow X$\;

\For{each layer $l \in \{1, 2, ..., L\}$}{
  // Linear transformation\;
  $z \leftarrow \mathbf{w}[l] \cdot h + b[l]$\;
  
  // Apply activation function\;
  $a \leftarrow \sigma_{\text{ReLU}}(z)$\;
  
  // Apply neuron gating (Eq.~\eqref{eq:tabdarts-gate})\;
  $h \leftarrow a \odot \sigma_{\text{GS-STE}}(\mathbf{g}[l])$\;
}

// Compute loss\;
$\mathcal{L} \leftarrow \text{loss\_fn}(h, Y)$\;

\Return $\mathcal{L}$
  
\caption{{\sc Forward Pass}}
\label{algo:forward}
\end{algorithm}

\section{Datasets}
\label{sec:data}
We evaluate \method~using six tabular datasets: four telecom-specific datasets representative of telecom use cases of well-known prediction tasks and two generic large tabular datasets for comprehensive benchmarking. Detailed information about these datasets is provided in the following subsections and Table~\ref{tab:datasets}, respectively.

\subsection{Received Signal Strength (RSS) Prediction} Three datasets address the challenge of predicting received signal strength (Reference Signal Received Power (RSRP) or path loss) across multiple frequency bands using measurements from a single primary carrier. Each dataset represents distinct network configurations, and frequency scenarios generated using different simulators. Across all datasets, the prediction task follows a consistent framework: given a UE location, signal strength metrics from the primary carriers across all cells are used to predict the signal strength for the secondary carriers. As an example, for DeepMIMO, given 18 cells in the network each with one primary and three secondary carriers there are $18\times1=18$ input features and $18\times3=54$ prediction tasks. Prediction tasks are formed as a regression problem. These results can be used to predict the strongest cell on the secondary carrier. This task has previously been studied by Masood et al.~\cite{farooq_multi-task_2024}.

\subsubsection{DeepMIMO} The dataset\footnote{\texttt{https://www.deepmimo.net/}} originates from a deployment scenario where we predict path loss for high-frequency secondary carriers (28 GHz, 60 GHz, and 140 GHz) using primary carriers (3.5 GHz) path loss measurements. It uses the Outdoor Urban Microcellular scenario to simulate 18 base stations and approximately 31,000 unique UE locations.

\subsubsection{Sim-A} The dataset models a cellular layout with three base stations with three cells each. It includes 100 simulation snapshots, each containing 1,000 randomly placed UEs, totalling 100,000 UE locations. In this task, we use RSRP measurements from an LTE carrier at 900 MHz to predict RSRP for a secondary NR carrier at 4.5 GHz.

\subsubsection{Sim-B} The dataset simulates a network with three base stations, each with three cells. It includes 50,000 UE location samples generated using 100 distinct UE identities. Here, the RSS task involves predicting RSRP values for secondary carriers at 1800 MHz, 2600 MHz, and 3500 MHz based on primary carrier RSRP at 800 MHz.

\subsection{Service Performance Prediction}
The fourth telecom dataset focuses on video-on-demand (VoD) frame rate (FR) prediction. It consists of server-side resource utilisation metrics; such as, disk I/O statistics, network statistics, CPU core utilisation, memory, and swap space utilisation, from a cluster of nine high-performance servers offering VoD services. The cluster receives a load according to a sinusoid function ranging 20{--}120 clients/minute generated using a Poisson process.
The goal is to predict the frame rate experienced by end users based on backend resource metrics, providing a proxy for user experience monitoring in service assurance tasks. The task is a regression problem.
For further information, we refer to Yanggratoke et al.~\cite{yanggratoke_predicting_2015}.

\subsection{Non-telecom Tabular Datasets}
As we are proposing a general tabular \ac{NAS} method, we additionally evaluate our method on two large-scale generic tabular datasets that are commonly used in the deep learning literature: CoverType, which involves forest cover type classification, and Higgs, which contains simulation data for predicting the Higgs boson particle in high-energy physics experiments.\footnote{Both datasets were downloaded from \texttt{openml.org}.}

\begin{table}[h]
  \caption{Summary of datasets.}
  \label{tab:datasets}
  \centering
  \begin{tabular}{lccc}

    \toprule
    {\textbf{Dataset}} 
    & \textbf{Task type} & \textbf{Input / Output} & {\textbf{Samples}}\\
    \midrule
    VoD (FR)                    & Regression              & 46 / \phantom{0}1                   & \phantom{0}50,000      \\
    DeepMIMO (RSS)              & Regression              & 18 / 54                             & \phantom{0}31,419      \\
    Sim-A (RSS)                 & Regression              & \phantom{0}9 / \phantom{0}9         & \phantom{0}99,999      \\
    Sim-B (RSS)                 & Regression              & \phantom{0}9 / 27                   & \phantom{0}50,100      \\
    \midrule
    CoverType                   &  Classification           & 54 / 7                                  &  581,012                \\
    Higgs                       & Classification           & 24 / 2                                  & 940,160                \\
    \bottomrule
    
  \end{tabular}
\end{table}

\section{Experiments} 
\label{sec:eval}

In this section, we analyse \method~to demonstrate its behaviour and the reasoning behind choices we have made during its development. Further, since we present a new tabular \ac{NAS} method, we compare it against the two existing tabular \ac{NAS} methods --- AgE and TabNAS.
Since \method~also functions as a training technique that achieves high accuracy, we have included a comparison with a large \ac{MLP} configured to represent the largest possible architecture within our search space. We do these evaluations on four different telecom datasets, presented in Section~\ref{sec:data}. Additionally, since our method can be applied to tabular data more generally, we evaluated it on two more datasets that are frequently used when evaluating \ac{ML} models for tabular data. We begin the section with details on the implementation of the methods and the experiments to ensure reproducibility.

\subsection{Implementation Details}
\label{sec:details}
\subsubsection*{Experimental setup}
To ensure a fair comparison between \ac{NAS} methods, we standardised the search space and evaluation procedures across all approaches and datasets. The architecture search is constrained to \acp{MLP} with five layers and a maximum width of 512 units per layer. This configuration was chosen as we observed diminishing returns in performance beyond this size. AgE and TabNAS require a predefined set of allowable layer widths. For these methods, we have selected 20 discrete options between 2 and 512. In contrast, our method does not require a predefined list and is free to select any width between 1 and 512. We did not extend the search spaces of AgE and TabNAS to include all possible widths as doing so would further increase their already substantial computational cost.
All methods train and evaluate selected architectures using a learning rate of $0.001$, with a maximum of 300 epochs and early stopping with a patience of 20. GPU experiments were conducted on an NVIDIA H100 for the RSS datasets and an NVIDIA A30 for the remaining datasets.

AgE and TabNAS select an architecture as a result of the search process. We report the final test performance of these architectures by creating new randomly initialised models and training them until convergence. \method, on the other hand, provides both an architecture and its weights. Therefore, we are not required to retrain the architecture from random initialisation. Instead, we warm start a new model from these weights and train for a small number of additional epochs.

\subsubsection*{\method}
While tuning hyperparameters for each dataset individually could yield better results, we opt for simplicity and use the same hyperparameters for \method~across all datasets. We set $\tau=1$, the architecture learning rate of $0.05$, and initialise our gate parameters to $-3$, giving them approximately 4.7\% probability of being open --- starting the search from small architectures.

Traditional DARTS methods often use ``unrolling'' to look ahead to the next architecture state when training the weights of the models. We have only seen increase in computation time when using unrolling, therefore do not use it in \method.

\subsubsection*{Other NAS methods}
For our comparative study, we use the \texttt{DeepHyper}\footnote{\texttt{https://github.com/deephyper/deephyper}} implementation of AgE. TabNAS does not have public code; therefore, we implemented it ourselves in PyTorch same as \method.
One advantage of AgE is its ability to parallelise computation over multiple devices. We were not able to parallelise over many GPUs; therefore, the experiments have been done in parallel on 20 CPU cores. This does not change the method and, therefore, its predictive performance. To ensure a fair comparison with other \ac{NAS} methods with regard to search time, we report all time measurements for CPUs as well.
For TabNAS, we have set the number of pre-training epochs to be one-fourth of the total epochs, as described by the authors. While a key feature of TabNAS is its ability to constrain the maximum number of parameters in the architecture, we do not utilise this feature in our comparisons to ensure that all methods are evaluated on the same search space. We use 2048 Monte-Carlo samples, a 0.001 RL learning rate and a momentum of 0.9. For AgE, we use a population of 100, a sample size of 10 and 300 iterations.



\subsection{Analysis of \method}

\begin{figure}
    \centering
    \includegraphics[width=0.85\columnwidth]{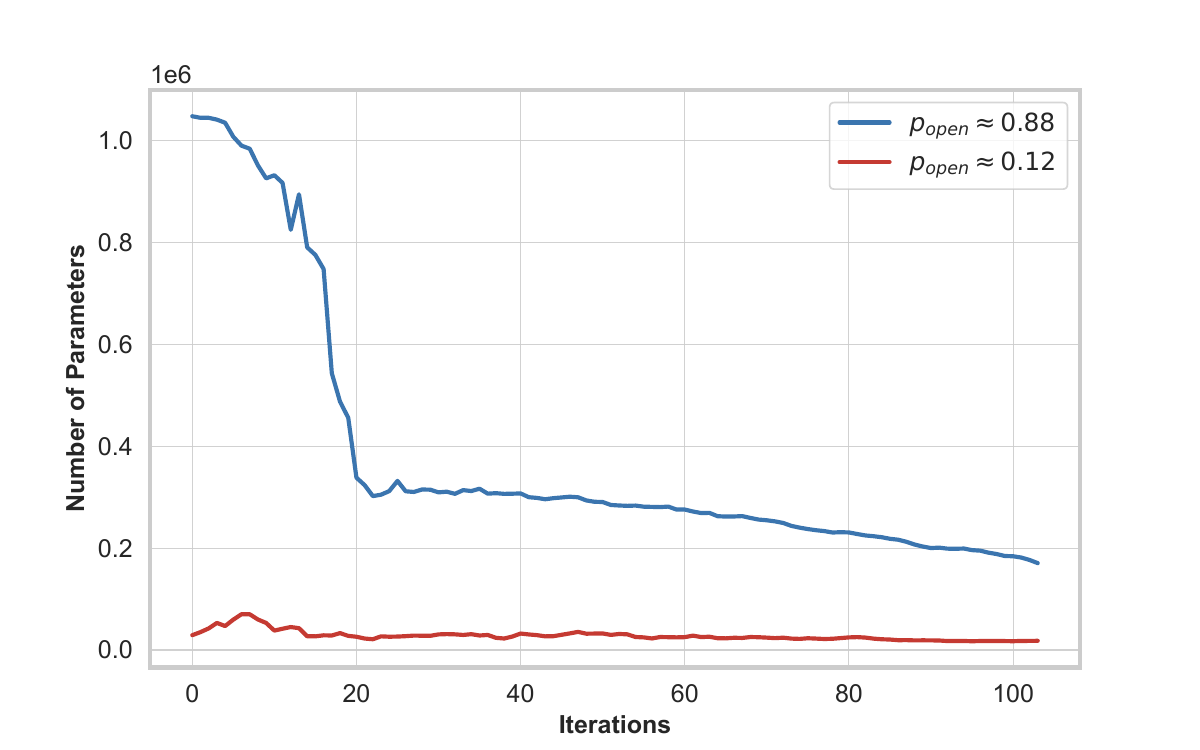}
    \caption{Size of the architecture during a search process given initialisation of gates with low/high probability of being open, for the VoD dataset. Initialisation of the gates determines the starting size of the architectures and consequentially the size of the final architecture. }
    \label{fig:gate-params-overtime}
\end{figure}

Our experiments highlight the significant influence of gate initialisation on \ac{NAS} outcomes.
We found that initialising the gate probabilities to be mostly open or mostly closed has an impact on the size of the final architecture. As seen in Fig.~\ref{fig:gate-params-overtime}, initialising all gates to be open with a small probability ($p_{\text{open}} \approx 12\%$) starts the architecture search from a small size and slowly grows it, while initialising to a large probability ($p_{\text{open}} \approx 88\%$) starts from a large architecture and shrinks it. 
\method~uses early stopping based on validation loss, terminating the search when validation performance stops improving. This causes the final architecture sizes to be biased towards their starting architecture size.

\begin{figure}
    \centering
    \includegraphics[width=0.95\columnwidth]{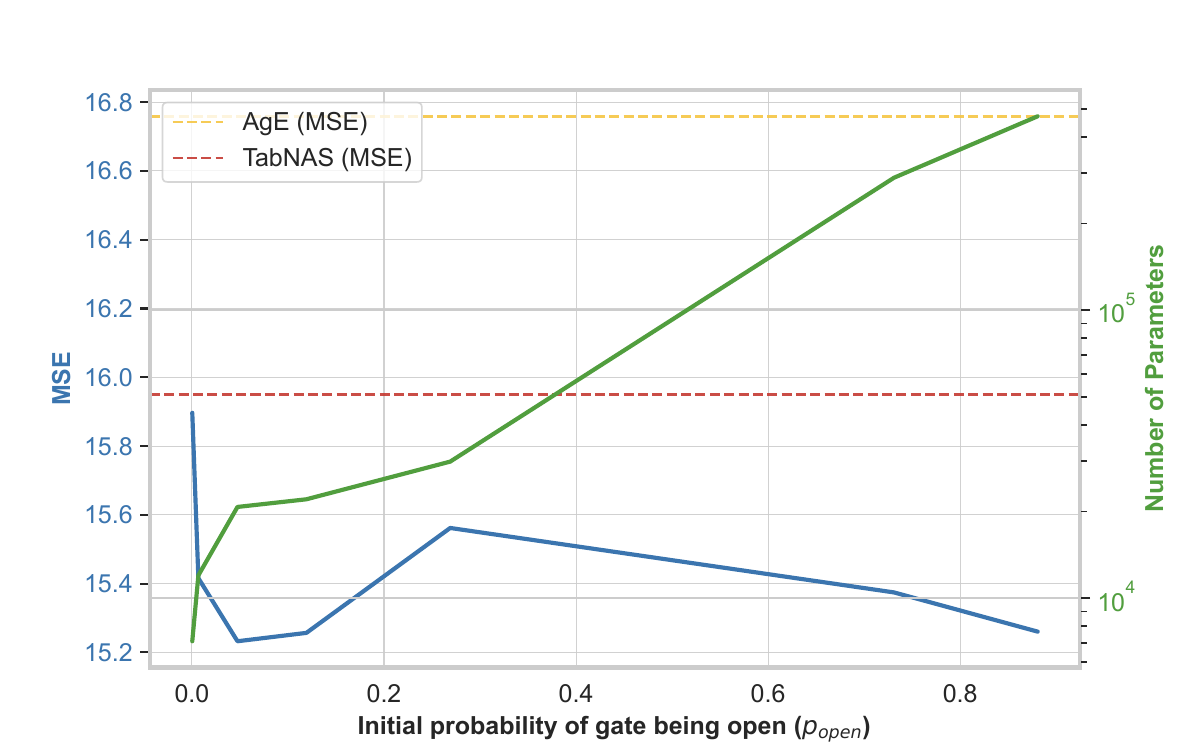}
    \caption{Architecture size and prediction error given initial gate state on VoD dataset. With low initial gate probability, \method~is able to find small architectures with low error.}
    \label{fig:gate-param-mse}
\end{figure}

More comprehensive experimentation in Fig.~\ref{fig:gate-param-mse} provides the final architecture size and test error for different neural architecture searches with different gate initialisations. Increasing the initial probability of gates being open results in a larger final architecture. Furthermore, we see that there isn't a substantial difference in the test error of these different initialisations. Specifically, all tested initialisations outperform other \ac{NAS} methods.
There is, however, a preferred range for the initialisation where we can obtain an architecture that both achieves low error while having a low number of parameters --- in this case, $p_{\text{open}} \sim 4 - 12\%$.

\subsection{Comparative Results}

\begin{table*} 
\centering
\begin{threeparttable}
\caption{Comparison of \method~with a large MLP and other NAS methods.}
\scriptsize
\begin{tabular}{lllcccc}
\toprule
\multirow{2}{*}{\textbf{Type}} & \multirow{2}{*}{\textbf{Dataset}} & \multirow{2}{*}{\textbf{Metric}} & {\textbf{Naive}} & \multicolumn{3}{c}{\textbf{NAS Method}} \\
\cmidrule(rl){4-4}
\cmidrule(rl){5-7}
& & & Large MLP & TabNAS & AgE & \method~(Ours) \\
\midrule
\multirow{8}{*}{\textbf{Telecom}} & \multirow{2}{*}{\textbf{VoD (FR)}} & MSE$\downarrow$ & \textbf{15.477 $\pm$ 0.417} & \phantom{0}15.949 $\pm$ \phantom{0}0.456 & \phantom{0}16.759 $\pm$ \phantom{00}3.539 & \textbf{\phantom{0}15.246 $\pm$ \phantom{0}0.622} \\
 & & \# Parameters$\downarrow$ & 1,048,576 & \phantom{0}56,576 $\pm$ 41,616 & 182,886 $\pm$ 207,455 & \textbf{\phantom{0}21,071 $\pm$ \phantom{0}5,018} \\
\cmidrule{2-7}
 & \multirow{2}{*}{\textbf{DeepMIMO (RSS)}} & MSE$\downarrow$ & 34.070 $\pm$ 1.981 & \phantom{0}81.593 $\pm$ 64.113 & \phantom{0}36.026 $\pm$ \phantom{00}1.411 & \textbf{\phantom{0}30.555 $\pm$ \phantom{0}0.367} \\
 & & \# Parameters$\downarrow$ & 1,048,576 & \phantom{0}63,590 $\pm$ 54,593\tnote{1} & 575,488 $\pm$ 170,563 & \textbf{141,337 $\pm$ 16,378} \\
\cmidrule{2-7}
 & \multirow{2}{*}{\textbf{Sim-A (RSS)}} & MSE$\downarrow$ & \textbf{20.297 $\pm$ 1.582} & \phantom{0}57.426 $\pm$ 69.821 & \textbf{\phantom{0}20.579 $\pm$ \phantom{00}1.781} & \textbf{\phantom{0}20.719 $\pm$ \phantom{0}0.270} \\
 & & \# Parameters$\downarrow$ & 1,048,576 & 110,526 $\pm$ 97,567\tnote{1} & 489,882 $\pm$ 103,872 & \textbf{\phantom{0}86,929 $\pm$ 11,567} \\
\cmidrule{2-7}
 & \multirow{2}{*}{\textbf{Sim-B (RSS)}} & MSE$\downarrow$ & \phantom{0}\textbf{7.692 $\pm$ 0.735} & \phantom{0}68.359 $\pm$ 53.491 & \phantom{00}9.372 $\pm$ \phantom{00}0.978 & \textbf{\phantom{00}7.666 $\pm$ \phantom{0}0.197} \\
 & & \# Parameters$\downarrow$ & 1,048,576 & \phantom{0}60,736 $\pm$ 65,297\tnote{1} & 470,357 $\pm$ 205,914 & \textbf{\phantom{0}89,615 $\pm$ \phantom{0}9,372} \\
\midrule
\multirow{4}{*}{\textbf{Non-Telecom}} & \multirow{2}{*}{\textbf{Covertype}} & Accuracy$\uparrow$ & 95.353 $\pm$ 0.193 & \phantom{0}94.691 $\pm$ \phantom{0}0.668 & \textbf{\phantom{0}95.528 $\pm$ \phantom{00}0.161} & \phantom{0}95.241 $\pm$ \phantom{0}0.192 \\
 & & \# Parameters$\downarrow$ & 1,048,576 & 142,131 $\pm$ 73,451 & 529,429 $\pm$ 121,975 & \textbf{\phantom{0}69,313 $\pm$ \phantom{0}4,784} \\
\cmidrule{2-7}
 & \multirow{2}{*}{\textbf{Higgs}} & Accuracy$\uparrow$ & 74.303 $\pm$ 0.148 & \phantom{0}74.603 $\pm$ \phantom{0}0.245 & \phantom{0}69.822 $\pm$ \phantom{0}10.459 & \textbf{\phantom{0}75.090 $\pm$ \phantom{0}0.139} \\
 & & \# Parameters$\downarrow$ & 1,048,576 & 138,525 $\pm$ 82,722 & \phantom{0}19,354 $\pm$ \phantom{00}9,944\tnote{1} & \textbf{\phantom{0}30,198 $\pm$ \phantom{0}5,176} \\
 \bottomrule
\end{tabular}
\label{tab:res-acc}
\begin{tablenotes}
\item[1] Parameter count ignored from bolding due to its significant detriment to prediction performance.
\item[$\downarrow$$\uparrow$] Downward ($\downarrow$) and upward ($\uparrow$) pointing arrows show whether smaller or larger values are better respectively.
\end{tablenotes}
\end{threeparttable}
\end{table*} 


As we have previously discussed, there are three dimensions of optimisation that are important for autonomously designing neural network architectures for telecom use cases, namely prediction error, architecture size, and search time. For a comprehensive evaluation, we compare \method~on all the datasets against state-of-the-art tabular \ac{NAS} methods and a large \ac{MLP}. We use a large \ac{MLP} as a naive baseline model that is the largest architecture in the search space and, therefore, much larger than anything the \ac{NAS} methods end up selecting.
We report prediction performance (mean-squared error or accuracy depending on the task) and final model size as the number of parameters (weights) in the selected architectures in Table~\ref{tab:res-acc}. Bold formatting indicates the best results that are statistically significant based on the t-test ($p < 0.05$).

Previous tabular \ac{NAS} methods manage to find architectures that are smaller than the large \ac{MLP}; however, they cannot improve the predictive performance. \method, on the other hand, can find architectures that match or exceed large \ac{MLP} prediction performance while containing 1-2 orders of magnitudes fewer parameters. Both TabNAS and AgE experience situations where they either don't converge or can't find a good architecture, causing a high variance in results.  Compared to previous \ac{NAS} methods, \method~not only achieves lower error for most datasets but also achieves more consistently good results (lower standard deviation). Architectures selected by \method~are also smaller than the ones selected by previous tabular \ac{NAS} methods. This makes them more suitable for many telecom use cases where there are limited computational resources or requirements on low inference latency.

\begin{table} 
\centering
\begin{threeparttable}
\caption{Comparison of search times (wall-clock time in hours) for different methods on a CPU and a GPU.}
\scriptsize
\begin{tabular}{llccc}
\toprule
\multirow{2}{*}{\textbf{Dataset}} & \multirow{2}{*}{\textbf{Device}} & \multicolumn{3}{c}{\textbf{Method}} \\
\cmidrule(l){3-5}
& & {TabNAS} &  {AgE\tnote{1}} & {\method~(Ours)} \\
\midrule
\multirow{2}{*}{\textbf{VoD}} & CPU & \phantom{0}1.32 $\pm$ 0.43 & \phantom{00}9.95 $\pm$ \phantom{0}4.20 & \textbf{\phantom{0}0.81 $\pm$ \phantom{0}0.36} \\
 & GPU & \phantom{0}1.46 $\pm$ 0.86 & \phantom{0}{--} & \textbf{\phantom{0}0.04 $\pm$ \phantom{0}0.01} \\
\midrule
\multirow{2}{*}{\textbf{DeepMIMO}} & CPU & \phantom{0}\textbf{0.76 $\pm$ 0.27} & \phantom{0}31.98 $\pm$ 16.02 & \phantom{0}1.26 $\pm$ \phantom{0}0.39 \\
 & GPU & \phantom{0}0.69 $\pm$ 0.26 & \phantom{0}{--} & \textbf{\phantom{0}0.07 $\pm$ \phantom{0}0.01} \\
\midrule
\multirow{2}{*}{\textbf{Sim-A}} & CPU & \phantom{0}4.51 $\pm$ 1.10 & \phantom{0}59.28 $\pm$ 18.94 & \textbf{\phantom{0}1.72 $\pm$ \phantom{0}0.62} \\
 & GPU & \phantom{0}1.80 $\pm$ 0.86 & \phantom{0}{--} & \textbf{\phantom{0}0.15 $\pm$ \phantom{0}0.05} \\
\midrule
\multirow{2}{*}{\textbf{Sim-B}} & CPU & \phantom{0}\textbf{1.55 $\pm$ 0.38} & \phantom{0}41.76 $\pm$ 23.27 & \textbf{\phantom{0}1.57 $\pm$ \phantom{0}0.45} \\
 & GPU & \phantom{0}0.97 $\pm$ 0.45 & \phantom{0}{--} & \textbf{\phantom{0}0.08 $\pm$ \phantom{0}0.02} \\
\midrule
\multirow{2}{*}{\textbf{Covertype}} & CPU & \phantom{0}\textbf{7.72 $\pm$ 2.64} & 232.84 $\pm$ 82.88 & 21.40 $\pm$ \phantom{0}6.71 \\
 & GPU & \phantom{0}8.06 $\pm$ 1.80 & \phantom{0}{--} & \textbf{\phantom{0}1.52 $\pm$ \phantom{0}0.11} \\
\midrule
\multirow{2}{*}{\textbf{Higgs}} & CPU & \textbf{13.51 $\pm$ 6.65} & 106.39 $\pm$ 23.19 & \textbf{22.37 $\pm$ 15.08} \\
 & GPU & 16.88 $\pm$ 5.39 & \phantom{0}{--} & \textbf{\phantom{0}2.10 $\pm$ \phantom{0}0.53} \\
\bottomrule
\end{tabular}
\label{tab:res-time}
\begin{tablenotes}
\item[1] AgE was run on 20 parallel CPU cores. No parallelisation for \method~and TabNAS.
\end{tablenotes}
\end{threeparttable}
\end{table} 

Search time measurement for the \ac{NAS} methods showcase the advantage of using gradient descent compared to \ac{RL} and \ac{ES}, especially while running on GPUs (see Table~\ref{tab:res-time}). \ac{RL} and \ac{ES} based \ac{NAS} methods require training and evaluating many architectures to be able to learn from them --- they are sample inefficient. Gradient-based methods, on the other hand, are able to learn by optimising the architectures given the loss landscape. This process is, therefore, much more efficient than the alternatives. This is especially true when utilising GPUs due to gradient-based optimisations' inherent parallelisability.
Our experiments show that evolutionary search, AgE, is significantly slower than the other two methods due to its requirement of completely training each architecture that is picked during the search, even though it uses 20 CPU cores to run the process in parallel. Speed comparisons of TabNAS and \method~on GPUs clearly show the advantage of gradient-based optimisation versus comparatively inefficient RL-based optimisation.
Furthermore, since \method~trains the architecture during the search, there is no requirement to retrain the final architecture. Instead, we can use the final weights of the method and extract the weights corresponding to the selected architecture. We use these weights to warm-start the architecture and train for a few additional epochs, which achieves an even better performance.

It is worth noting that although the range of possible architectures in the search spaces of the selected NAS methods is the same, because \method~has a more granular control over the architecture the number of possible choices it has to pick from is larger.
As described in Section~\ref{sec:details}, TabNAS and AgE can choose out of 20 options per layer ($20^{5} = 3,2 \cdot 10^6$ unique architectures), while \method~chooses an architecture out of 512 different widths per layer ($512^5 \approx 3,5 \cdot 10^{13}$ unique architectures).

In conclusion, \method~enables starting the search from small architectures and growing them to select architectures that are smaller than state-of-the-art tabular \ac{NAS} methods while having a better prediction performance. \method~is able to converge to a final architecture much faster than previous methods. Additionally, there is no need to retrain the final selected architecture since the weights of the architecture are optimised in conjunction with the architecture search.

\section{Conclusion} 
\label{sec:conclusion}

We address the challenge of automating neural network design for telecom data, where the manual process of architecture selection has been time‑consuming and often suboptimal due to the need to balance prediction performance with resource constraints.
We introduce TabGNS, the first gradient-based architecture–search approach tailored for tabular data. Our evaluations show that TabGNS delivers superior performance across three dimensions: it matches or exceeds state-of-the-art predictive accuracy on 5 out of 6 datasets, while requiring only $18\text{--}49\%$ of the parameters used by prior methods, and achieving a $5\text{--}36\times$ reduction in search time. By automating the architecture–search process, TabGNS eliminates the bottlenecks of manual design, enabling faster retraining cycles and rapid adaptation to evolving network conditions. The resulting architectures exhibit a substantially smaller footprint, reducing hardware demands and minimising inference latency, a critical factor for time-sensitive telecommunication tasks. Finally, TabGNS attains these efficiency gains while simultaneously improving predictive accuracy.


Our future work will extend the evaluation to cover more datasets within network management and telecom and explore the applicability of gradient-based gated neuron selection to other types of layers and architectures, such as convolutional or recurrent layers.

\section*{Acknowledgement}
This work was partially supported by the Wallenberg AI, Autonomous Systems and Software Program (WASP) funded by the Knut and Alice Wallenberg Foundation. We would like to thank Dinand Roeland at Ericsson Research, for his constructive feedback which greatly improved the clarity and quality of the work.

\bibliographystyle{IEEEtran}
\bibliography{references}

\end{document}